\documentclass{article}
\usepackage{amssymb}
\usepackage{graphicx} 
\usepackage{float} 
\usepackage{multirow}
\usepackage[table,xcdraw]{xcolor}

\usepackage[preprint]{corl_2025} 

\title{VLH: Vision-Language-Haptics Foundation Model}

\author {The list of authors has been anonymized for submission
\thanks{Anonymized for submission}}

\author{
\footnotesize
\begin{tabular}{cc}
Luis Francisco Moreno Fuentes & Muhammad Haris Khan \\
Intelligent Space Robotics Laboratory & Intelligent Space Robotics Laboratory \\
Skolkovo Institute of Science and Technology & Skolkovo Institute of Science and Technology \\
Russian Federation & Russian Federation \\
\texttt{Luis.Moreno@skoltech.ru} & \texttt{Haris.Khan@skoltech.ru} \\[0.5cm]
Miguel Altamirano Cabrera & Valerii Serpiva \\
Intelligent Space Robotics Laboratory & Intelligent Space Robotics Laboratory \\
Skolkovo Institute of Science and Technology & Skolkovo Institute of Science and Technology \\
Russian Federation & Russian Federation \\
\texttt{m.altamirano@skoltech.ru} & \texttt{Valerii.Serpiva@skoltech.ru} \\[0.5cm]
Dmitri Iarchuk & Yara Mahmoud \\
Intelligent Space Robotics Laboratory & Intelligent Space Robotics Laboratory \\
Skolkovo Institute of Science and Technology & Skolkovo Institute of Science and Technology \\
Russian Federation & Russian Federation \\
\texttt{Dmitrii.Iarchuk@skoltech.ru} & \texttt{Yara.Mahmoud@skoltech.ru} \\[0.5cm]
Issatay Tokmurziyev & Dzmitry Tsetserukou \\
Intelligent Space Robotics Laboratory & Intelligent Space Robotics Laboratory \\
Skolkovo Institute of Science and Technology & Skolkovo Institute of Science and Technology \\
Russian Federation & Russian Federation \\
\texttt{Issatay.Tokmurziyev@skoltech.ru} & \texttt{D.Tsetserukou@skoltech.ru} \\[0.5cm]
\end{tabular}
}

\begin{document}
\maketitle 

\begin{figure}[H]
    \centering
    \includegraphics[width=0.9\textwidth]{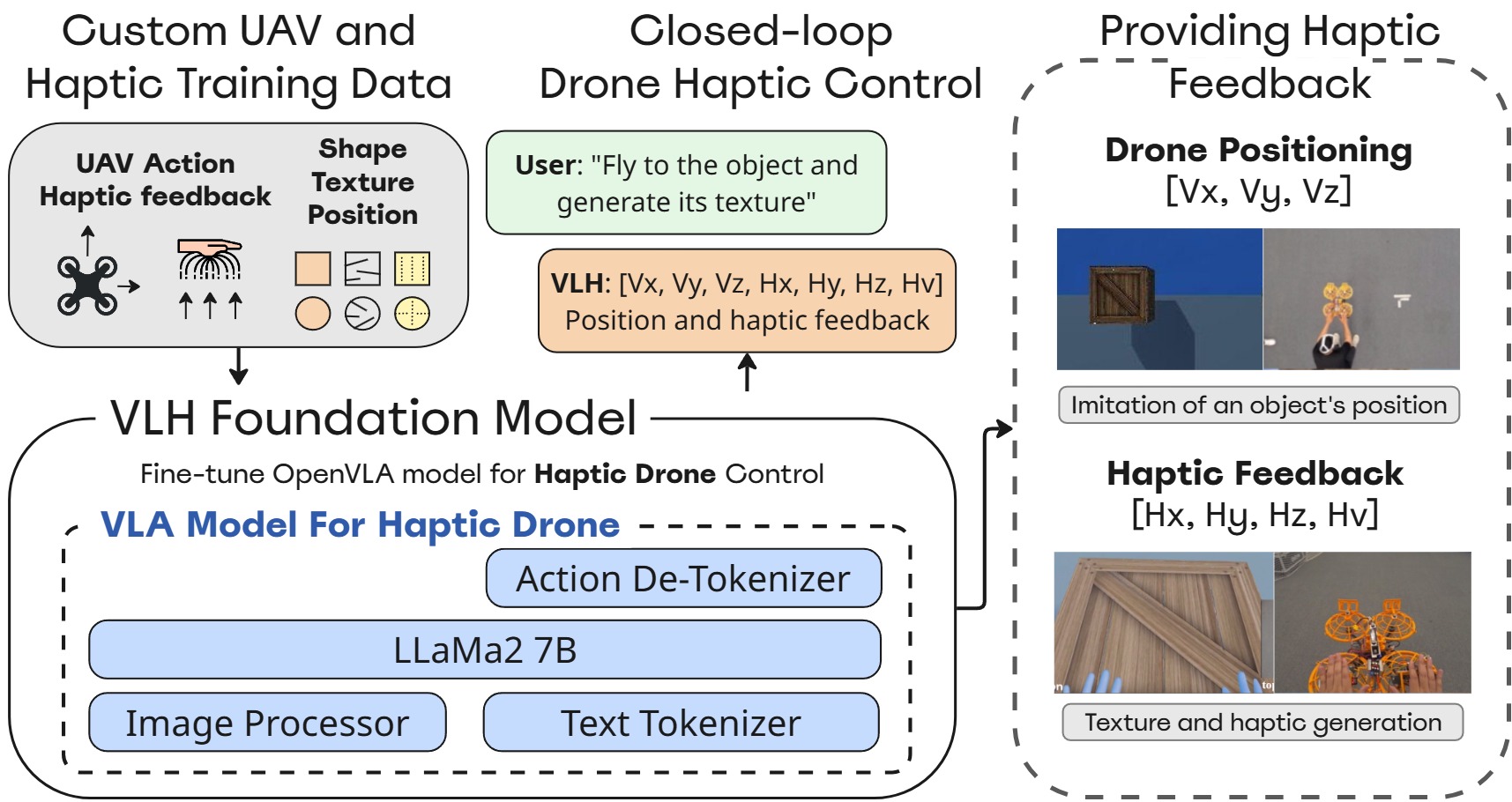}
    \caption{VLH is the first model specifically designed for the HapticDrone system. It processes video streams from VR and top-down drone views with language commands to generate velocity actions (Vx, Vy, Vz) and haptic feedback (Hx, Hy, Hz) as well as vibration feedback (Hv). At the core of the system is an aerial haptic platform with two five-bar linkage arrays, designed to deliver mid-air, localized force and vibration cues. VLH combines mechanical actuation with perceptual reasoning by using visual data to estimate object geometry and surface properties.}
    \label{fig:teaser}
\end{figure}


\begin{abstract}
We present VLH, a novel Visual-Language-Haptic Foundation Model that unifies perception, language, and tactile feedback in aerial robotics and virtual reality. Unlike prior work that treats haptics as a secondary, reactive channel, VLH synthesizes mid-air force and vibration cues as a direct consequence of contextual visual understanding and natural language commands. Our platform comprises an 8-inch quadcopter equipped with dual inverse five-bar linkage arrays for localized haptic actuation, an egocentric VR camera, and an exocentric top-down view. Visual inputs and language instructions are processed by a fine-tuned OpenVLA backbone—adapted via LoRA on a bespoke dataset of 450 multimodal scenarios—to output a 7-dimensional action vector $(V_x, V_y, V_z, H_x, H_y, H_z, H_v)$. INT8 quantization and a high-performance server ensure real-time operation at 4–5 Hz. In human-robot interaction experiments (90 flights), VLH achieved a 56.7\% success rate for target acquisition (mean reach time 21.3 s, pose error 0.24 m) and 100\% accuracy in texture discrimination. Generalization tests yielded 70.0\% (visual), 54.4\% (motion), 40.0\% (physical), and 35.0\% (semantic) performance on novel tasks. These results demonstrate VLH’s ability to co-evolve haptic feedback with perceptual reasoning and intent, advancing expressive, immersive human-robot interactions.


\end{abstract}



\section{Introduction}
Touch is a fundamental aspect of human communication, conveying emotions, intentions, and physical interactions with unparalleled nuance. Yet, in VR and robotics, haptic technologies remain limited to static, pre-programmed cues, failing to capture the dynamic, context-sensitive nature of touch. This limitation hinders the development of immersive human-robot interactions, where touch could serve as a powerful communicative channel.

Current haptic systems, such as vibration-based actuators or glove-based feedback, often provide reactive responses that lack expressiveness. Recent studies, however, suggest that haptics can transcend these limitations. For example, sparse vibration patterns paired with language models can convey emotional cues like happiness or sadness  \cite{10.5555/3721488.3721720}, enhancing user engagement in VR. Similarly, wearable devices like Dextres \cite{DextrES} offer multimodal haptic feedback in teleoperation, improving task performance by synchronizing tactile and visual inputs, but their on-body form factor constrains free-space interaction. In social robotics, affective touch is recognized as a communicative act that carries emotional intent \cite{affectivetouch2024}, underscoring the potential for haptics to become an expressive medium.

Advancements in generative models and aerial robotics further underscore the promise of context-aware, multimodal interfaces. HapticGen—a text-to-vibration model—automatically generates vibrotactile signals from textual descriptions \cite{sung2025hapticgen}, while Haptic Repurposing with GenAI transforms everyday objects into adaptive haptic interfaces via generative AI \cite{wang2024haptic}. In virtual reality,  VRHapticDrones leverages quadcopters as passive touch proxies, providing fixed contact surfaces when users intersect virtual objects \cite{hoppe2018vrhapticdrones}, whereas VLH actively synthesizes force and vibration patterns in concert with flight commands.  Similarly, vision language action (VLA) frameworks like RaceVLA \cite{serpiva2025racevlavlabasedracingdrone} and CognitiveDrone \cite{lykov2025cognitivedronevlamodelevaluation} excel at language-conditioned UAV navigation but omit tactile modalities.

Despite these advancements, a critical gap remains: haptic systems rarely function as co-evolving, expressive modalities integrated with real-time perception and user intent, particularly in dynamic environments like aerial robotics. Addressing this gap is essential for creating intuitive, emotionally rich interactions between humans and machines, enabling safer and more effective operations in immersive settings.

We propose VLH: Visual-Language-Haptic Foundation Model, a pioneering framework that redefines haptics as a generative, expressive modality seamlessly integrated with vision and language in aerial robotics. By combining inverse five-bar linkage arrays on an aerial platform with a dual-vision system (egocentric from the VR user and exocentric from an external camera), VLH generates real-time, context-aware haptic feedback. This positions touch as a communicative channel that conveys physical sensations and shared understanding.

Our contributions include: (1) a novel framework treating haptics as a generative outcome of perception and intent, (2) an aerial haptic platform, for localized feedback, (3) a dual-vision system for enhanced perceptual reasoning, (4) an adapted VLA model for UAV control and haptic response, and (5) empirical validation through real-world experiments demonstrating 57\% positional alignment. VLH paves the way for a new class of virtual experiences that are physically grounded, emotionally rich, and cognitively intuitive, advancing human-robot interaction in immersive environments.

\section{Related Work}

Recent studies in immersive technologies have tried to unify perception, interaction, and embodiment. While considerable progress has been made in computer vision and language understanding, the integration of touch—as a generative, expressive modality—remains underexplored. We situate our work at the intersection of generative haptics, vision-language-action modeling, and aerial physical interfaces, and highlight prior research that laid the groundwork for VLH.

Haptic feedback has long been used to enhance immersion in virtual and remote interactions. Traditional approaches include wearable gloves, vibration actuators, or exoskeletons, often designed to simulate predefined contact sensations \cite{hong2023vibration}. Other research has focused on integrating haptic feedback into wearable systems to improve user interaction with unmanned aerial vehicles (UAV). For instance, a study introduced a wearable vibrotactile feedback system that provides directional obstacle information to UAV operators, enhancing collision avoidance capabilities \cite{AeroHaptix}. Additionally, research has explored the role of haptic feedback in social interactions within virtual environments, demonstrating its effectiveness as a non-verbal communication tool to enhance coordination and the sense of agency during shared virtual experiences \cite{HapticSensing}.

The development of VLA \cite{brohan2023rt}, \cite{vuong2023open},\cite{khan2025evolution} models has progressed towards creating general-purpose robotic systems capable of understanding and executing complex tasks. OpenVLA, an open-source VLA model, has been trained on a diverse collection of real-world robot demonstrations, showcasing strong results in generalist manipulation tasks\cite{kim2024openvlaopensourcevisionlanguageactionmodel}.Furthermore, Helix represents a generalist VLA model that unifies perception, language understanding, and learned control, enabling humanoid robots to perform a variety of tasks with enhanced adaptability \cite{Helix}.

Innovations in aerial haptics have led to the development of systems like VRHapticDrones \cite{HapticDrone}, which utilize quadcopters as levitating haptic feedback proxies. These drones provide touchable surfaces that can render haptic feedback in virtual reality environments, offering a novel approach to untethered, spatially dynamic feedback.  The OmniRace \cite{omnirace6dhandpose} project demonstrates effective hand-based drone control for human-robot interaction but lacks haptic feedback, limiting user awareness and reducing control precision during high-speed tasks. Additionally, research has proposed omnidirectional aerial robots equipped with haptic feedback mechanisms to enhance telepresence experiences, providing users with intuitive control interfaces \cite{mellet2024designcontrolomnidirectionalaerial}.

The integration of haptic feedback into VR simulations has been shown to enhance learning experiences by providing realistic physical interactions. For example, immersive procedural training in VR, incorporating both visual and haptic cues, has been systematically reviewed to assess its effectiveness in replicating real-life training scenarios\cite{Immersivetraining}. Moreover, studies have investigated the role of embodiment in VR learning, finding that increased levels of immersion and interactivity positively affect learning outcomes \cite{klingenberg2024does}. In summary, prior systems either emphasize haptic feedback without leveraging language-based decision-making, or leverage vision-language models without tactile interaction. This gap motivates our VLH framework: unlike previous work, VLH combines generative haptic rendering with language-guided control in aerial robots.  This integration of generative haptics with VLA reasoning for UAVs has not been addressed in existing literature, making VLH a novel contribution to multimodal robotics.
	

\section{System Overview}

\subsection{System Architecture}

The VLH system comprises four primary components: a dual vision system, a VLA model, an intent inference and haptic mapping module, and an aerial haptic interface. The dual vision system includes an egocentric camera, capturing the user's first-person perspective within the VR environment, and an exocentric camera, providing an external viewpoint of the user and their surroundings. This combination offers comprehensive visual data, enabling accurate interpretation of user actions and environmental context.


The developed hardware system is an 8-inch custom-built drone equipped with a SpeedyBee F405 flight controller running ArduPilot firmware, interfaced with the Robot Operating System (ROS) through MAVROS. An onboard OrangePi 5B computer manages high-level control tasks, software packages, and communication with a high-performance computing (HPC) server hosting the VLH model via a Flask API. Precise localization is achieved using a Vicon motion capture system with 14 infrared cameras, providing accurate real-time tracking of the drone's position and movement. The OrangePi 5B further controls a haptic feedback device via an I2C connection to PWM PCA9685 drivers, actuating lightweight DMS44 and high-torque HS-70MG servomotors. Six invert five-bar linkage mechanisms are connected to these servomotors, and located on the drone as two arrays, enabling effective tactile force feedback to the operator, thus enhancing user immersion during interaction with VR environments. 

\subsection{The VLH model}

\begin{figure}[t]
\vspace{+3mm}
  \centering
  \includegraphics[width=1.0\textwidth]{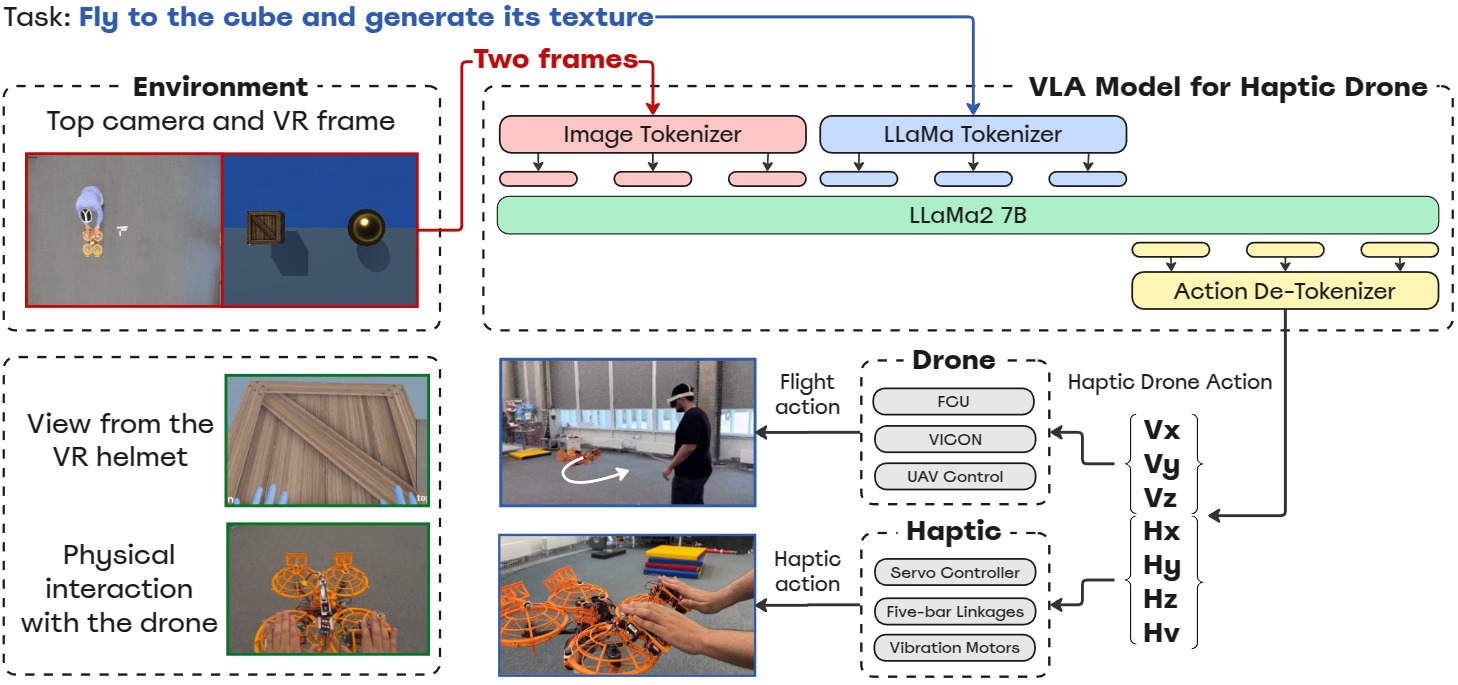}
  \caption{VLH and Haptic Drone system architecture.}
  \label{fig:system_overwiew}
  \vspace{-0.4cm}
\end{figure}

The VLH system (shown in Fig. \ref{fig:system_overwiew}) is a fine-tuned VLA model specifically adapted for UAV flight control and haptic device feedback. Building upon the OpenVLA model developed by Stanford AI Lab, we fine-tuned it using a custom dataset containing drone flight sequences paired with detailed haptic feedback corresponding to various textures, shapes, and surface properties of virtual objects. Each data sample links drone trajectories to VR object positions while simultaneously capturing associated haptic responses. While our system architecture retains the core multimodal input-output structure of OpenVLA, we adapted it to meet the specific demands of flight and haptic interaction. Inputs to the model include two synchronized top-down frames: one from the real-world flight camera and another from the VR environment, both aligned in position and orientation, along with a natural language command. The model outputs a refined 7D action vector designed for aerial haptic platform operation: three components for linear velocity control $(V_x, V_y, V_z)$ and four for haptic feedback — three for directional force outputs $(H_x, H_y, H_z)$ and one for vibration intensity $(H_{v})$.


Unlike traditional navigation strategies that rely on predefined waypoints, the VLH model implements a dynamic, frame‐by‐frame control loop. Upon receiving an instruction, the drone continuously processes incoming visual frames to produce immediate control actions, enabling real‐time trajectory adjustments without awaiting waypoint completion. This approach affords highly responsive navigation within dynamic VR settings. Haptic feedback is learned by correlating VR images with surface features: the model infers object shape, size, and texture from each frame and commands a custom delta haptic device accordingly. To facilitate real‐time performance, the VLA model is quantized to INT8, substantially reducing inference latency. Deployed on a high‐performance server with an NVIDIA RTX 4090 GPU and optimized API pipelines, the system sustains a 4–5 Hz update rate, balancing responsiveness, computational efficiency, and precise delivery of both haptic sensations and flight controls.

\section{Dataset Collection and Traning Pipline}

For dataset collection, a top-down view of real‐world interactions with the flying haptic interface and of the VR environment—where users manipulated virtual objects—was captured. Multiple flight sessions were conducted under varying initial conditions (e.g., start positions and object placements), yielding 50 drone‐flight trajectory sets with corresponding action commands \((V_x,V_y,V_z)\), recorded via a Vicon motion‐capture system. Each session comprised on average 110 frames from a top‐view camera during autonomous flight. Real‐world data were aligned with VR frames featuring three object shapes (cube, sphere, cone) and three texture categories (food, plastic, other), each paired with haptic signals \((H_x,H_y,H_z,H_v)\), resulting in 450 unique visual‐physical‐action combinations. All VR and real‐world images were resized to \(640\times320\) pixels to comply with the VLH model’s input requirements.

The dataset was structured in accordance with the \emph{Reinforcement Learning Dataset Specification} (RLDS) to ensure compatibility with OpenVLA’s modules for actions, haptic signals, visual inputs, and textual directives. It was employed to fine-tune the 7 billion-parameter OpenVLA-7b model using a Low-Rank Adaptation (LoRA) scheme (rank 32), thereby updating only a minimal subset of weights to conserve memory and enhance efficiency. Training proceeded over 4 000 steps with a batch size of 16 and a learning rate of \(5\times10^{-4}\), executed on a single NVIDIA A100 GPU to accommodate the multimodal data demands.


\section{Results}

\subsection{Haptic Pattern Recognition Study}

To assess the effectiveness of the proposed haptic feedback module to generate  recognizable tactile patterns, a user study was conducted. Three different object shapes were rendered with both inverse five-bar linkage arrays (cube, sphere, and cone) with three levels of vibration (high, low, and no vibration). Consequently, a total of nine unique tactile patterns were generated and presented to the participants for evaluation.

Twelve participants (8 males, 4 females, aged 22–35 years, mean 25.8 $\pm$3.7) completed the study. After providing informed consent, participants underwent a training session to familiarize themselves with the patterns. Each pattern was rendered two times during training, and a visual reference of the patterns was provided throughout the session.

\begin{figure}[t]
  \centering
  \includegraphics[width=0.5\textwidth]{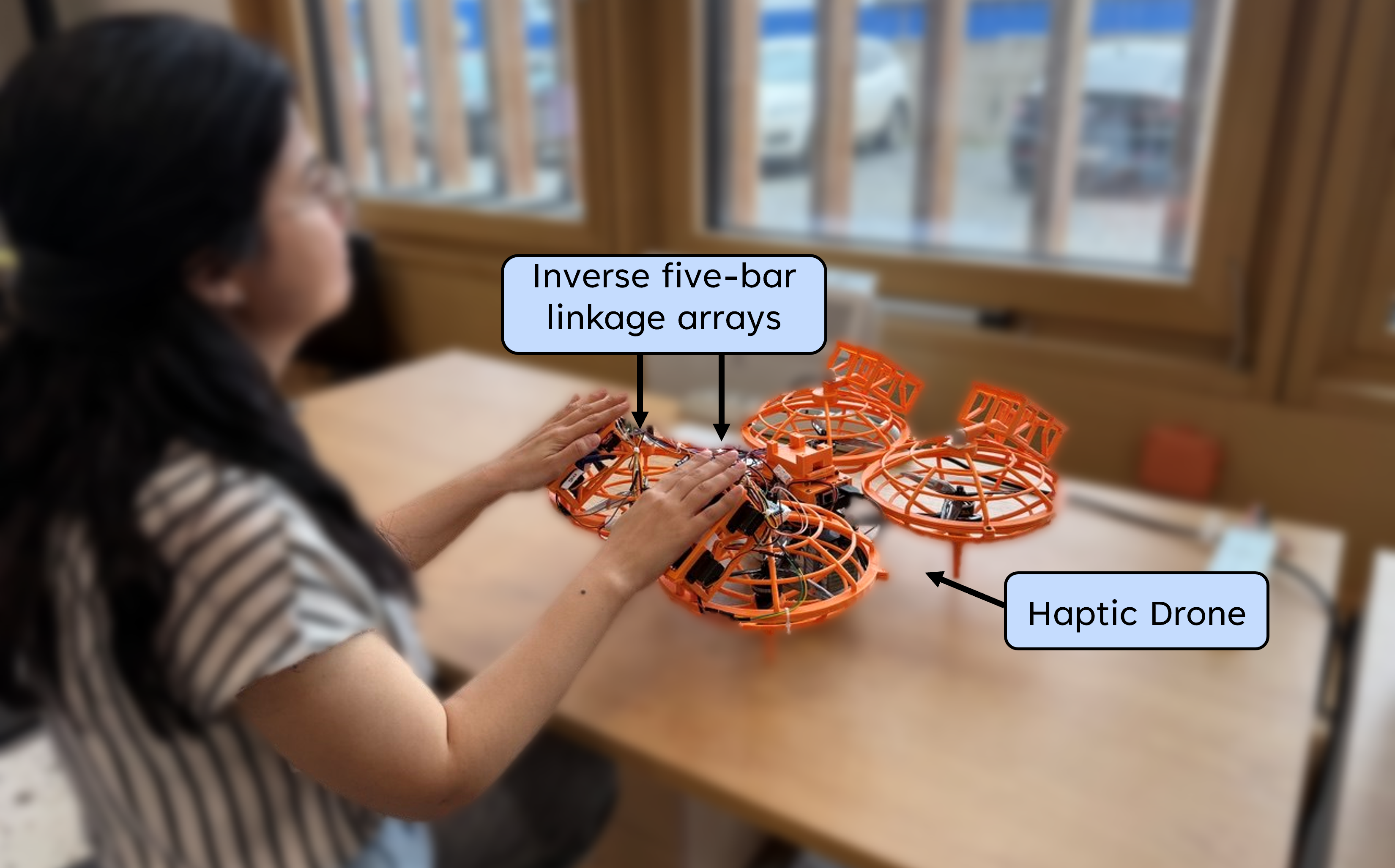}
  \caption{Participant seated at a desk during the evaluation. The users interacted with both hands with the inverse five-bar linkage arrays.}
  \label{fig:user_study}
  \vspace{-0.4cm}
\end{figure}

During the evaluation, participants were asked to sit in front of a desk and to locate their hands on the device, as shown in Fig.~\ref{fig:user_study}. The experimenter used a graphical user interface (GUI) on a PC to select the patterns that the users perceived. Each of the 9 patterns (three shapes, three vibration conditions) was presented three times in random order, resulting in 27 trials per participant. Participants provided feedback on the perceived sensations at the end of the study.

Recognition of shapes averaged 80.7\%, and recognition of the levels of vibration averaged 81.3\%, and the combination of the shapes and level of vibration, the recognition rate averaged 57\%. A confusion matrix summarizing the results is shown in Table \ref{table:confusion}.


\begin{table}[]

\centering{
\caption{Confusion Matrix for Actual and Perceived Pattern Recognition.}
\label{table:confusion}

\begin{tabular}{|ccl|ccccccccc|}
\hline

\multicolumn{3}{|c|}{}                                                                                  & \multicolumn{9}{c|}{Answers   (Predicted Class)}                                                                                                                                                                                                                                                                                                                                                                                                                                                                                                                                                                                                     \\ \cline{4-12} 
\multicolumn{3}{|c|}{}                                                                                  & \multicolumn{3}{c|}{circle}                                                                                                                                                                                                    & \multicolumn{3}{c|}{square}                                                                                                                                                                             & \multicolumn{3}{c|}{cone}                                                                                                                                                                                 \\ \cline{4-12} 
\multicolumn{3}{|c|}{\multirow{-3}{*}{\%}}                                                              & \multicolumn{1}{c|}{high}                                                & \multicolumn{1}{c|}{low}                                                 & \multicolumn{1}{c|}{null}                                                & \multicolumn{1}{c|}{high}                         & \multicolumn{1}{c|}{low}                                                 & \multicolumn{1}{c|}{null}                                                & \multicolumn{1}{c|}{high}                                                & \multicolumn{1}{c|}{low}                                                 & null                                                \\ \hline
\multicolumn{1}{|c|}{}                           & \multicolumn{1}{c|}{}                         & high & \multicolumn{1}{c|}{\cellcolor[HTML]{506A76}{\color[HTML]{FFFFFF} 0.50}} & \multicolumn{1}{c|}{\cellcolor[HTML]{A8B5BB}0.25}                        & \multicolumn{1}{c|}{\cellcolor[HTML]{FFFFFF}-}                        & \multicolumn{1}{c|}{\cellcolor[HTML]{CFD6D9}0.14} & \multicolumn{1}{c|}{\cellcolor[HTML]{F6F7F8}0.03}                        & \multicolumn{1}{c|}{\cellcolor[HTML]{FFFFFF}-}                        & \multicolumn{1}{c|}{\cellcolor[HTML]{E2E7E9}0.08}                        & \multicolumn{1}{c|}{\cellcolor[HTML]{FFFFFF}-}                        & \cellcolor[HTML]{FFFFFF}-                        \\ \cline{3-12} 
\multicolumn{1}{|c|}{}                           & \multicolumn{1}{c|}{}                         & low  & \multicolumn{1}{c|}{\cellcolor[HTML]{CFD6D9}0.14}                        & \multicolumn{1}{c|}{\cellcolor[HTML]{3C5A67}{\color[HTML]{FFFFFF} 0.56}} & \multicolumn{1}{c|}{\cellcolor[HTML]{FFFFFF}-}                        & \multicolumn{1}{c|}{\cellcolor[HTML]{E2E7E9}0.08} & \multicolumn{1}{c|}{\cellcolor[HTML]{CFD6D9}0.14}                        & \multicolumn{1}{c|}{\cellcolor[HTML]{FFFFFF}-}                        & \multicolumn{1}{c|}{\cellcolor[HTML]{ECEFF0}0.06}                        & \multicolumn{1}{c|}{\cellcolor[HTML]{F6F7F8}0.03}                        & \cellcolor[HTML]{FFFFFF}-                        \\ \cline{3-12} 
\multicolumn{1}{|c|}{}                           & \multicolumn{1}{c|}{\multirow{-3}{*}{circle}} & null & \multicolumn{1}{c|}{\cellcolor[HTML]{FFFFFF}-}                        & \multicolumn{1}{c|}{\cellcolor[HTML]{E2E7E9}0.08}                        & \multicolumn{1}{c|}{\cellcolor[HTML]{46626E}{\color[HTML]{FFFFFF} 0.53}} & \multicolumn{1}{c|}{\cellcolor[HTML]{FFFFFF}-} & \multicolumn{1}{c|}{\cellcolor[HTML]{F6F7F8}0.03}                        & \multicolumn{1}{c|}{\cellcolor[HTML]{D8DEE1}0.11}                        & \multicolumn{1}{c|}{\cellcolor[HTML]{FFFFFF}-}                        & \multicolumn{1}{c|}{\cellcolor[HTML]{ECEFF0}0.06}                        & \cellcolor[HTML]{BBC6CA}0.19                        \\ \cline{2-12} 
\multicolumn{1}{|c|}{}                           & \multicolumn{1}{c|}{}                         & high & \multicolumn{1}{c|}{\cellcolor[HTML]{E2E7E9}0.08}                        & \multicolumn{1}{c|}{\cellcolor[HTML]{D8DEE1}0.11}                        & \multicolumn{1}{c|}{\cellcolor[HTML]{FFFFFF}-}                        & \multicolumn{1}{c|}{\cellcolor[HTML]{81949C}0.36} & \multicolumn{1}{c|}{\cellcolor[HTML]{6D838D}0.42}                        & \multicolumn{1}{c|}{\cellcolor[HTML]{F6F7F8}0.03}                        & \multicolumn{1}{c|}{\cellcolor[HTML]{FFFFFF}-}                        & \multicolumn{1}{c|}{\cellcolor[HTML]{FFFFFF}-}                        & \cellcolor[HTML]{FFFFFF}-                        \\ \cline{3-12} 
\multicolumn{1}{|c|}{}                           & \multicolumn{1}{c|}{}                         & low  & \multicolumn{1}{c|}{\cellcolor[HTML]{F6F7F8}0.03}                        & \multicolumn{1}{c|}{\cellcolor[HTML]{D8DEE1}0.11}                        & \multicolumn{1}{c|}{\cellcolor[HTML]{FFFFFF}-}                        & \multicolumn{1}{c|}{\cellcolor[HTML]{C5CED2}0.17} & \multicolumn{1}{c|}{\cellcolor[HTML]{1F4150}{\color[HTML]{FFFFFF} 0.64}} & \multicolumn{1}{c|}{\cellcolor[HTML]{ECEFF0}0.06}                        & \multicolumn{1}{c|}{\cellcolor[HTML]{FFFFFF}-}                        & \multicolumn{1}{c|}{\cellcolor[HTML]{FFFFFF}-}                        & \cellcolor[HTML]{FFFFFF}-                        \\ \cline{3-12} 
\multicolumn{1}{|c|}{}                           & \multicolumn{1}{c|}{\multirow{-3}{*}{square}} & null & \multicolumn{1}{c|}{\cellcolor[HTML]{FFFFFF}-}                        & \multicolumn{1}{c|}{\cellcolor[HTML]{F6F7F8}0.03}                        & \multicolumn{1}{c|}{\cellcolor[HTML]{BBC6CA}0.19}                        & \multicolumn{1}{c|}{\cellcolor[HTML]{FFFFFF}-} & \multicolumn{1}{c|}{\cellcolor[HTML]{ECEFF0}0.06}                        & \multicolumn{1}{c|}{\cellcolor[HTML]{153948}{\color[HTML]{FFFFFF} 0.67}} & \multicolumn{1}{c|}{\cellcolor[HTML]{FFFFFF}-}                        & \multicolumn{1}{c|}{\cellcolor[HTML]{FFFFFF}-}                        & \cellcolor[HTML]{ECEFF0}0.06                        \\ \cline{2-12} 
\multicolumn{1}{|c|}{}                           & \multicolumn{1}{c|}{}                         & high & \multicolumn{1}{c|}{\cellcolor[HTML]{E2E7E9}0.08}                        & \multicolumn{1}{c|}{\cellcolor[HTML]{ECEFF0}0.06}                        & \multicolumn{1}{c|}{\cellcolor[HTML]{FFFFFF}-}                        & \multicolumn{1}{c|}{\cellcolor[HTML]{ECEFF0}0.06} & \multicolumn{1}{c|}{\cellcolor[HTML]{FFFFFF}-}                        & \multicolumn{1}{c|}{\cellcolor[HTML]{FFFFFF}-}                        & \multicolumn{1}{c|}{\cellcolor[HTML]{1F4150}{\color[HTML]{FFFFFF} 0.64}} & \multicolumn{1}{c|}{\cellcolor[HTML]{C5CED2}0.17}                        & \cellcolor[HTML]{FFFFFF}-                        \\ \cline{3-12} 
\multicolumn{1}{|c|}{}                           & \multicolumn{1}{c|}{}                         & low  & \multicolumn{1}{c|}{\cellcolor[HTML]{ECEFF0}0.06}                        & \multicolumn{1}{c|}{\cellcolor[HTML]{D8DEE1}0.11}                        & \multicolumn{1}{c|}{\cellcolor[HTML]{FFFFFF}-}                        & \multicolumn{1}{c|}{\cellcolor[HTML]{FFFFFF}-} & \multicolumn{1}{c|}{\cellcolor[HTML]{E2E7E9}0.08}                        & \multicolumn{1}{c|}{\cellcolor[HTML]{ECEFF0}0.06}                        & \multicolumn{1}{c|}{\cellcolor[HTML]{C5CED2}0.17}                        & \multicolumn{1}{c|}{\cellcolor[HTML]{46626E}{\color[HTML]{FFFFFF} 0.53}} & \cellcolor[HTML]{FFFFFF}-                        \\ \cline{3-12} 
\multicolumn{1}{|c|}{\multirow{-9}{*}{\rotatebox{90}{Patterns}}} & \multicolumn{1}{c|}{\multirow{-3}{*}{cone}}   & null & \multicolumn{1}{c|}{\cellcolor[HTML]{FFFFFF}-}                        & \multicolumn{1}{c|}{\cellcolor[HTML]{F6F7F8}0.03}                        & \multicolumn{1}{c|}{\cellcolor[HTML]{CFD6D9}0.14}                        & \multicolumn{1}{c|}{\cellcolor[HTML]{FFFFFF}-} & \multicolumn{1}{c|}{\cellcolor[HTML]{FFFFFF}-}                        & \multicolumn{1}{c|}{\cellcolor[HTML]{ECEFF0}0.06}                        & \multicolumn{1}{c|}{\cellcolor[HTML]{FFFFFF}-}                        & \multicolumn{1}{c|}{\cellcolor[HTML]{E2E7E9}0.08}                        & \cellcolor[HTML]{0B3040}{\color[HTML]{FFFFFF} 0.69} \\ \hline
\end{tabular}}
\end{table}

\begin{table}[!htb]
    
    \begin{minipage}{.5\linewidth}
    \centering{
      \caption{Confusion Matrix for Actual and \\ Perceived Shapes.}
      
        \begin{tabular}{|cl|ccc|}
        
\hline
\multicolumn{2}{|c|}{}                                  & \multicolumn{3}{c|}{Answers   (Predicted Class)}                                                                                                                                                          \\ \cline{3-5} 
\multicolumn{2}{|c|}{\multirow{-2}{*}{}}                & \multicolumn{1}{l|}{circle}                                              & \multicolumn{1}{l|}{square}                                              & \multicolumn{1}{l|}{cone}                           \\ \hline
\multicolumn{1}{|c|}{}                         & circle & \multicolumn{1}{c|}{\cellcolor[HTML]{2E4E5C}{\color[HTML]{FFFFFF} 0.69}} & \multicolumn{1}{c|}{\cellcolor[HTML]{CED6D9}0.18}                        & \cellcolor[HTML]{DADFE2}0.14                        \\ \cline{2-5} 
\multicolumn{1}{|c|}{}                         & square & \multicolumn{1}{c|}{\cellcolor[HTML]{CBD3D7}0.19}                        & \multicolumn{1}{c|}{\cellcolor[HTML]{0B3040}{\color[HTML]{FFFFFF} 0.80}} & \cellcolor[HTML]{FFFFFF}0.02                        \\ \cline{2-5} 
\multicolumn{1}{|c|}{\multirow{-3}{*}{{Shapes}}} & cone   & \multicolumn{1}{c|}{\cellcolor[HTML]{D4DBDD}0.16}                        & \multicolumn{1}{c|}{\cellcolor[HTML]{EBEEF0}0.08}                        & \cellcolor[HTML]{173A4A}{\color[HTML]{FFFFFF} 0.76} \\ \hline
\end{tabular}}
    \end{minipage}%
    \begin{minipage}{.5\linewidth}
      \centering{
        \caption{    Confusion Matrix for Actual and \\ Perceived Levels of Vibration. }}
       \begin{tabular}{|cc|ccc|}
\hline
\multicolumn{2}{|c|}{}                                         & \multicolumn{3}{c|}{Answers (Predicted Class)}                                                                                                                                                          \\ \cline{3-5} 
\multicolumn{2}{|c|}{\multirow{-2}{*}{}}                       & \multicolumn{1}{c|}{high}                                                & \multicolumn{1}{c|}{low}                                                 & \multicolumn{1}{c|}{null}                           \\ \hline
\multicolumn{1}{|c|}{}                                  & high & \multicolumn{1}{c|}{\cellcolor[HTML]{4C6773}{\color[HTML]{FFFFFF} 0.65}} & \multicolumn{1}{c|}{\cellcolor[HTML]{A0AFB5}0.34}                        & \cellcolor[HTML]{FDFDFD}0.01                        \\ \cline{2-5} 
\multicolumn{1}{|c|}{}                                  & low  & \multicolumn{1}{c|}{\cellcolor[HTML]{BDC7CB}0.24}                        & \multicolumn{1}{c|}{\cellcolor[HTML]{355361}{\color[HTML]{FFFFFF} 0.73}} & \cellcolor[HTML]{F5F7F7}0.04                        \\ \cline{2-5} 
\multicolumn{1}{|c|}{\multirow{-3}{*}{\rotatebox{90}{Vibra}}} & null & \multicolumn{1}{c|}{\cellcolor[HTML]{FFFFFF}0.00}                        & \multicolumn{1}{c|}{\cellcolor[HTML]{DEE3E5}0.12}                        & \cellcolor[HTML]{0B3040}{\color[HTML]{FFFFFF} 0.88} \\ \hline
\end{tabular}
    \end{minipage} 
    \vspace{-0.4cm}
\end{table}
A two-way repeated measures ANOVA was conducted to assess the impact of the shape and the level of vibration on the recognition accuracy. The analysis revealed no statistically significant main effect for vibration (F(1, 68) = 1.074, p = 0.3034) or the shape (F(2, 68) = 0.537, p = 0.587). Furthermore, the interaction between vibration and temperature was not significant (F(2, 68) = 1.907, p = 0.156), indicating that recognition accuracy did not significantly differ across the various levels of vibration or the shape of the designed patterns.

\subsection{Evaluation of VLH Generalization}

\begin{figure}[t]
  \centering
  \includegraphics[width=.95\textwidth]{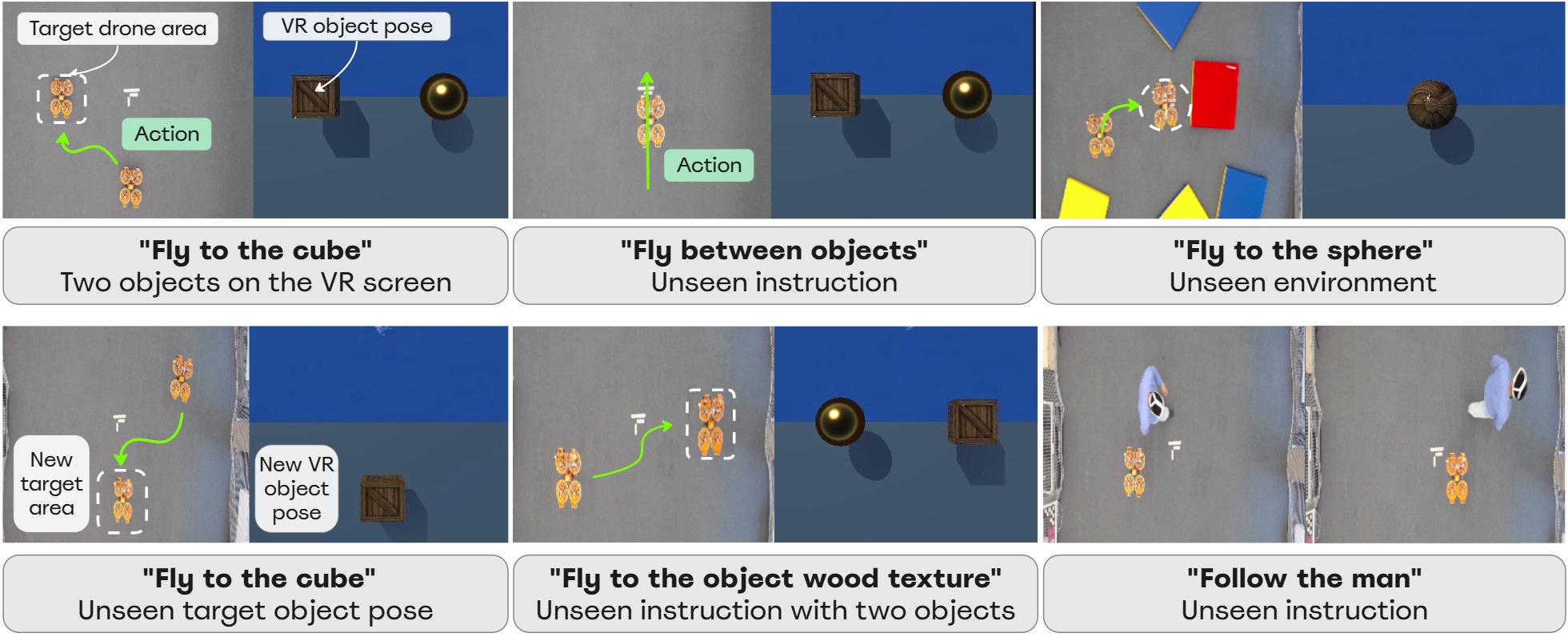}
  \caption{The figure illustrates the tasks used for generalization evaluation of VLH, tested across four new domains: visual, motion, physical, and semantic. These tasks assess VLH's adaptability to scenarios not included in the initial dataset, with the drone performing various tasks specified by the user.}
  \label{fig:new_tasks}
\end{figure}

To evaluate the generalization capabilities of VLH, we followed the experimental framework established in OpenVLA, adapting it to assess haptic feedback delivery using VR and the aerial haptic platform. The evaluation aimed to measure generalization across four distinct axes: visual, motion, physical, and semantic. These axes were tested through a comprehensive set of tasks, involving 90 experiments conducted with VLH, incorporating both flight and haptic reasoning. Samples of the task presented in Fig. \ref{fig:new_tasks}.

\begin{itemize}
    \item \textbf{Visual Generalization}: Tasks were designed to assess how effectively VLH adapts to new environments, featuring unseen backgrounds, objects, and variations in the appearance of target items.
   \item \textbf{Motion Generalization}: A varying number of objects were introduced in the same scene to evaluate VLH's ability to handle unseen object positions during dynamic motion.
  \item \textbf{Physical Generalization}: Tasks involved changes in the size and texture of VR objects, testing VLH’s ability to adjust to physical alterations in the environment.
  \item \textbf{Semantic Generalization}: Novel drone-specific tasks were introduced to evaluate the model’s capability to understand and respond to unfamiliar instructions and target objects.
\end{itemize}

\begin{figure}[t]
  \centering
  \includegraphics[width=0.75\textwidth]{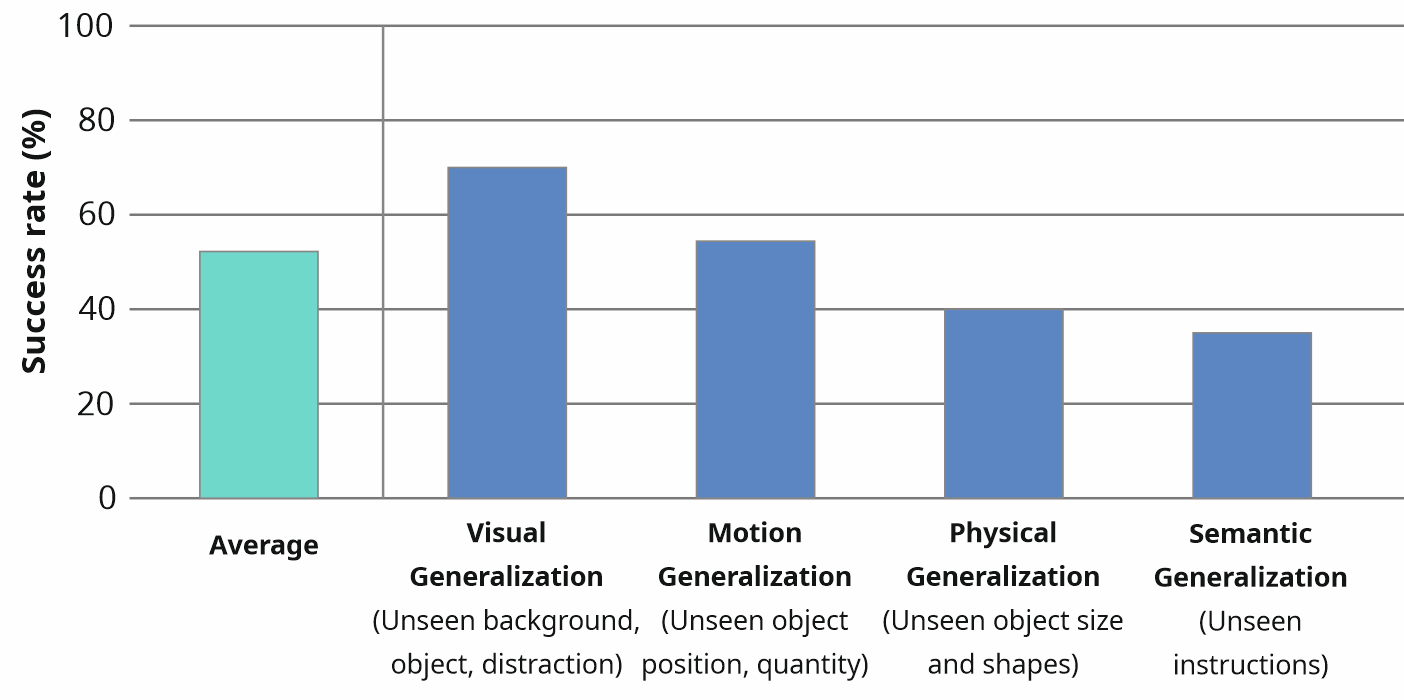}
  \caption{The figure illustrates the generalization evaluation of VLH, where it was tested on four new tasks: visual, motion, physical, and semantic.}
  \label{fig:succes_rate}
  \vspace{-0.4cm}
\end{figure}

The generalization performance of the VLH model across four axes — visual, motion, physical, and semantic — is summarized in Figure \ref{fig:succes_rate}, which presents both evaluation results and representative task examples. The VLH model achieved a visual generalization of 70.0\%, motion generalization of 54.4\%, physical generalization of 40.0\%, and semantic generalization of 35.0\%. These results demonstrate that VLH can effectively generalize across various task dimensions, particularly excelling in handling different visual appearances — even when tested in new environments or with distracting objects introduced into the scene, the model successfully maintained object tracking and texture prediction. In terms of motion variations, VLH showed robust performance by correctly recognizing shapes and generating appropriate drone actions even when additional objects, such as cubes and spheres, were simultaneously present in the scene. Although semantic generalization remains more challenging, with new instructions such as "follow the man" or recognizing and reacting to new textures presenting higher difficulty, overall performance highlights the model’s promising ability to adapt to novel instructions and dynamic environments, making it a strong candidate for real-world UAV applications requiring human-centered interaction.

\begin{figure}[t]
  \centering
  \includegraphics[width=1.0\textwidth]{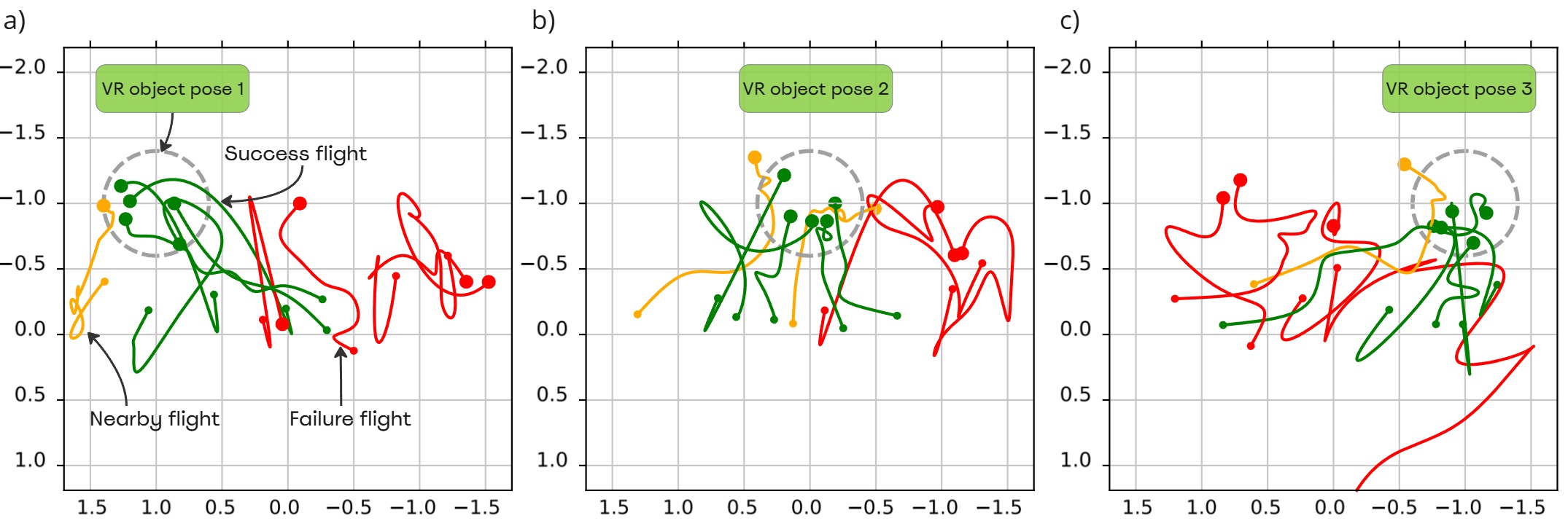}
  \caption{Visualization of drone flight trajectories during task evaluation. Plots (a), (b), and (c) show recorded flight paths toward three target poses. Plot (a) corresponds to the sphere at [1, -1] (radius 0.8 m), (b) to the target at [0, -1], and (c) to the target at [1, 1]. Green trajectories indicate successful flights with stable hovering for at least 5 sec, yellow denote partial success with shorter hovering, and red represent failed attempts where the drone missed the target.}
  \label{fig:exp_3_poses}
  \vspace{-0.2cm}
\end{figure}
\begin{figure}[t]
  \centering
  \includegraphics[width=0.75\textwidth]{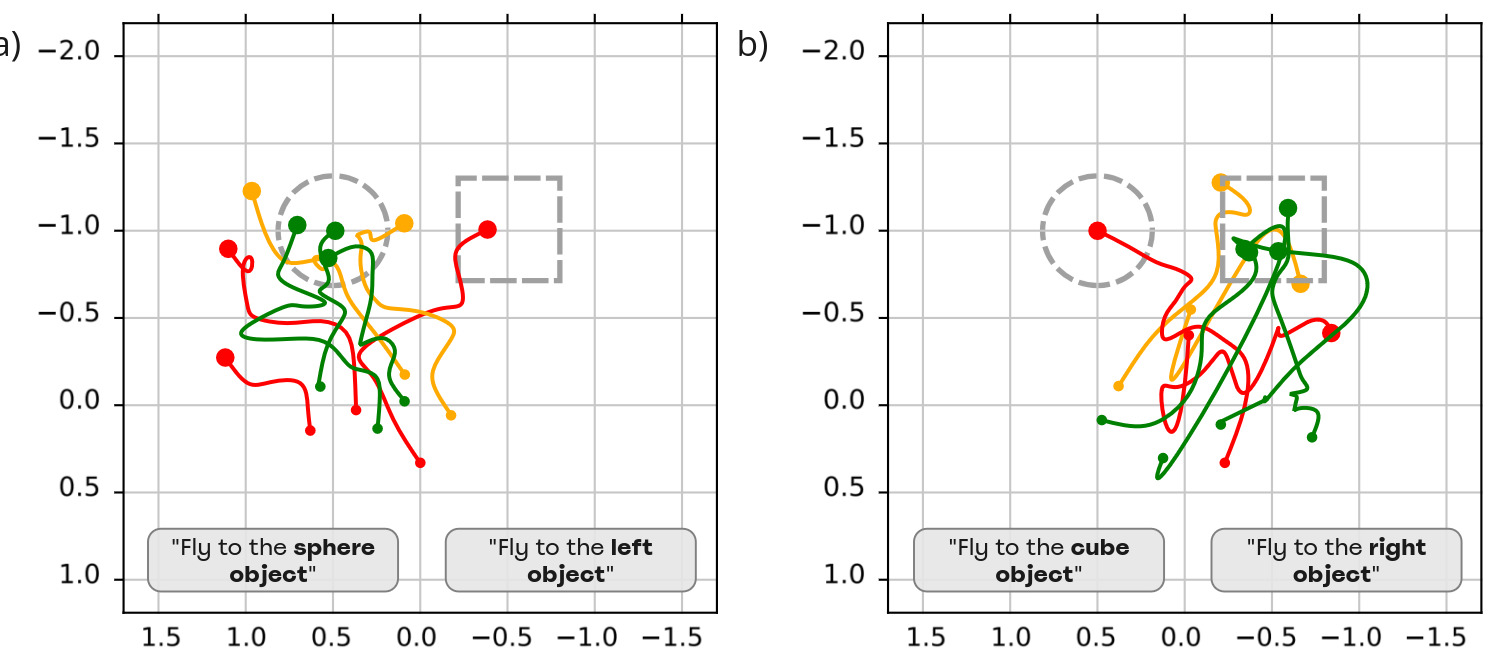}
  \caption{Recorded trajectories for the evaluation of system performance on unseen tasks with two objects present in the VR scene. (a) Flight toward the sphere (Fly to Sphere task) with an unseen command requiring flight to the left object. (b) Flight toward the cube (Fly to Cube task) with an unseen command requiring flight to the right object.}
  \label{fig:exp_2_objects}
  \vspace{-0.4cm}
\end{figure}

\subsection{VLH Evaluation of Flight Results}

To assess the system's performance in position following and target acquisition, we conducted multiple autonomous flights in which the system reached a target VR object and hovered within a specified threshold, allowing time for the operator to receive haptic feedback. Figure \ref{fig:exp_3_poses} (a, b, c) illustrates samples of the recorded trajectories in the task of reaching the VR object pose in three different positions and Figure \ref{fig:exp_3_poses} (d, e) represents the task of reaching a target object when two objects were spawned in the VR scene. The evaluation focused on two main criteria: success rate, defined as the percentage of flights that reached the target area within an acceptable threshold, and trajectory tracking performance, measured by the distance to the target and flight duration. The success rate across the different poses resulted in an overall success rate of 56.7\% across all trials. The average time to reach the target pose was 21.3 s, with a maximum solving time of 35.7 s and a minimum solving time of 11.9 s. During the hovering phase at the desired pose, the mean pose error was 0.24 m, with a standard error of 0.08 m.

\section{Conclusion}
The VLH (Visual-Language-Haptic) Foundation Model represents a pioneering advancement in human-robot interaction by seamlessly integrating visual perception, natural language processing, and haptic feedback within aerial robotics and virtual reality (VR) environments. Through its innovative aerial haptic platform, featuring an 8-inch quadcopter with dual inverse five-bar linkage arrays, and a dual-vision system combining egocentric and exocentric perspectives, VLH achieves context-aware, real-time haptic rendering and interaction. Experimental results demonstrate a 56.7\% success rate in target acquisition across 90 flights, with an average reach time of 21.3 seconds and a mean pose error of 0.24 meters, alongside 100\% accuracy in texture discrimination. The model’s generalization capabilities—70.0\% in visual, 54.4\% in motion, 40.0\% in physical, and 35.0\% in semantic domains—highlight its adaptability to diverse scenarios, though challenges remain in semantic and physical generalization. By treating haptics as a generative, expressive modality co-evolving with perception and intent, VLH sets a new benchmark for immersive, physically grounded, and emotionally rich interactions. This work lays a robust foundation for future developments in multimodal robotics, with significant potential to enhance VR applications in education, training, and human-robot collaboration. Future work will focus on integrating larger language models and expanding the dataset will strengthen semantic understanding, while leveraging generative models for realistic haptic rendering will address physical generalization challenges. Exploring multi-drone coordination and developing applications for educational and therapeutic VR environments will further unlock VLH’s transformative potential.
% ...

\clearpage



\begin{thebibliography}{21}
\providecommand{\natexlab}[1]{#1}
\providecommand{\url}[1]{\texttt{#1}}
\expandafter\ifx\csname urlstyle\endcsname\relax
  \providecommand{\doi}[1]{doi: #1}\else
  \providecommand{\doi}{doi: \begingroup \urlstyle{rm}\Url}\fi

\bibitem[Ren and Belpaeme(2025)]{10.5555/3721488.3721720}
Q.~Ren and T.~Belpaeme.
\newblock Touched by chatgpt: Using an llm to drive affective tactile interaction.
\newblock In \emph{Proc. of the 2025 ACM/IEEE Int. Conf. on Human-Robot Interaction}, HRI '25, page 1563–1567. IEEE Press, 2025.

\bibitem[Hinchet et~al.(2018)Hinchet, Vechev, Shea, and Hilliges]{DextrES}
R.~Hinchet, V.~Vechev, H.~Shea, and O.~Hilliges.
\newblock Dextres: Wearable haptic feedback for grasping in vr via a thin form-factor electrostatic brake.
\newblock In \emph{Proc. of the 31st Annual ACM Symposium on User Interface Software and Technology}, UIST '18, page 901–912, 2018.

\bibitem[McLaren et~al.(2024)McLaren, Gao, Yin, Reis~Guerra, Vyas, Morton, Cang, Chen, Sun, Li, Madden, and MacLean]{affectivetouch2024}
D.~McLaren, J.~Gao, X.~Yin, R.~Reis~Guerra, P.~Vyas, C.~Morton, X.~L. Cang, Y.~Chen, Y.~Sun, Y.~Li, J.~D.~W. Madden, and K.~E. MacLean.
\newblock What is affective touch made of? a soft capacitive sensor array reveals the interplay between shear, normal stress and individuality.
\newblock In \emph{Proc. of the 37th Annual ACM Symposium on User Interface Software and Technology}, UIST '24, 2024.

\bibitem[Sung et~al.(2025)Sung, John, Yoon, and Seifi]{sung2025hapticgen}
Y.~Sung, K.~John, S.~H. Yoon, and H.~Seifi.
\newblock Hapticgen: Generative text-to-vibration model for streamlining haptic design.
\newblock In \emph{Proc. of the 2025 CHI Conf. on Human Factors in Computing Systems}, pages 1--24, 2025.

\bibitem[Wang(2024)]{wang2024haptic}
H.~Wang.
\newblock Haptic repurposing with genai.
\newblock 2024.
\newblock arXiv:2406.07228.

\bibitem[Hoppe et~al.(2018)Hoppe, Knierim, Kosch, Funk, Futami, Schneegass, Henze, Schmidt, and Machulla]{hoppe2018vrhapticdrones}
M.~Hoppe, P.~Knierim, T.~Kosch, M.~Funk, L.~Futami, S.~Schneegass, N.~Henze, A.~Schmidt, and T.~Machulla.
\newblock Vrhapticdrones: Providing haptics in virtual reality through quadcopters.
\newblock In \emph{Proc. of the 17th Int. Conf. on Mobile and Ubiquitous Multimedia}, pages 7--18, 2018.

\bibitem[Serpiva et~al.(2025)Serpiva, Lykov, Myshlyaev, Khan, Abdulkarim, Sautenkov, and Tsetserukou]{serpiva2025racevlavlabasedracingdrone}
V.~Serpiva, A.~Lykov, A.~Myshlyaev, M.~H. Khan, A.~A. Abdulkarim, O.~Sautenkov, and D.~Tsetserukou.
\newblock Racevla: Vla-based racing drone navigation with human-like behaviour.
\newblock 2025.
\newblock arXiv:2503.02572.

\bibitem[Lykov et~al.(2025)Lykov, Serpiva, Khan, Sautenkov, Myshlyaev, Grik~Tadevosyan, and Tsetserukou]{lykov2025cognitivedronevlamodelevaluation}
A.~Lykov, V.~Serpiva, M.~H. Khan, O.~Sautenkov, A.~Myshlyaev, Y.~Y. Grik~Tadevosyan, and D.~Tsetserukou.
\newblock Cognitivedrone: A vla model and evaluation benchmark for real-time cognitive task solving and reasoning in uavs.
\newblock 2025.
\newblock arXiv:2503.01378.

\bibitem[Hong(2023)]{hong2023vibration}
S.~Hong.
\newblock Vibration-based wearable haptic feedback device and its applications.
\newblock \emph{Theoretical and Natural Science}, 17:\penalty0 104--109, 2023.

\bibitem[Huang et~al.(2025)Huang, Wang, Cheng, Ren, Cai, Valdivia, Mahadevan, and Wigdor]{AeroHaptix}
B.~Huang, Z.~Wang, Q.~Cheng, S.~Ren, H.~Cai, A.~A. Valdivia, K.~Mahadevan, and D.~Wigdor.
\newblock Aerohaptix: A wearable vibrotactile feedback system for enhancing collision avoidance in uav teleoperation.
\newblock \emph{IEEE Robotics and Automation Letters}, 10\penalty0 (5):\penalty0 4260--4267, 2025.

\bibitem[Shi and Shen(2024)]{HapticSensing}
Y.~Shi and G.~Shen.
\newblock Haptic sensing and feedback techniques toward virtual reality.
\newblock \emph{Research}, 7:\penalty0 0333, 2024.

\bibitem[Brohan et~al.(2023)Brohan, Brown, Carbajal, Chebotar, Chen, Choromanski, Ding, Driess, Dubey, Finn, et~al.]{brohan2023rt}
A.~Brohan, N.~Brown, J.~Carbajal, Y.~Chebotar, X.~Chen, K.~Choromanski, T.~Ding, D.~Driess, A.~Dubey, C.~Finn, et~al.
\newblock Rt-2: Vision-language-action models transfer web knowledge to robotic control.
\newblock 2023.
\newblock arXiv:2307.15818.

\bibitem[Vuong et~al.(2023)Vuong, Levine, Walke, Pertsch, Singh, Doshi, Xu, Luo, Tan, Shah, et~al.]{vuong2023open}
Q.~Vuong, S.~Levine, H.~R. Walke, K.~Pertsch, A.~Singh, R.~Doshi, C.~Xu, J.~Luo, L.~Tan, D.~Shah, et~al.
\newblock Open x-embodiment: Robotic learning datasets and rt-x models.
\newblock In \emph{Towards Generalist Robots: Learning Paradigms for Scalable Skill Acquisition@ CoRL2023}, 2023.

\bibitem[Khan et~al.(2025)Khan, Myshlyaev, Lykov, Cabrera, and Tsetserukou]{khan2025evolution}
M.~H. Khan, A.~Myshlyaev, A.~Lykov, M.~A. Cabrera, and D.~Tsetserukou.
\newblock Evolution 6.0: Evolving robotic capabilities through generative design.
\newblock 2025.
\newblock arXiv:2502.17034.

\bibitem[Kim et~al.(2024)Kim, Pertsch, Karamcheti, Xiao, Balakrishna, Nair, Rafailov, Foster, Lam, Sanketi, Vuong, Kollar, Burchfiel, Tedrake, Sadigh, Levine, Liang, and Finn]{kim2024openvlaopensourcevisionlanguageactionmodel}
M.~J. Kim, K.~Pertsch, S.~Karamcheti, T.~Xiao, A.~Balakrishna, S.~Nair, R.~Rafailov, E.~Foster, G.~Lam, P.~Sanketi, Q.~Vuong, T.~Kollar, B.~Burchfiel, R.~Tedrake, D.~Sadigh, S.~Levine, P.~Liang, and C.~Finn.
\newblock Openvla: An open-source vision-language-action model.
\newblock 2024.
\newblock arXiv:2406.09246.

\bibitem[Mei et~al.(2025)Mei, Zhuang, Miao, Yang, Jia, and Vinayak]{Helix}
Y.~Mei, Y.~Zhuang, X.~Miao, J.~Yang, Z.~Jia, and R.~Vinayak.
\newblock Helix: Serving large language models over heterogeneous gpus and network via max-flow.
\newblock In \emph{Proc. of the 30th ACM Int. Conf. on Architectural Support for Programming Languages and Operating Systems, Volume 1}, ASPLOS '25, page 586–602, 2025.

\bibitem[Hoppe et~al.(2018)Hoppe, Knierim, Kosch, Funk, Futami, Schneegass, Henze, Schmidt, and Machulla]{HapticDrone}
M.~Hoppe, P.~Knierim, T.~Kosch, M.~Funk, L.~Futami, S.~Schneegass, N.~Henze, A.~Schmidt, and T.~Machulla.
\newblock Vrhapticdrones: Providing haptics in virtual reality through quadcopters.
\newblock In \emph{Proc. of the 17th Int. Conf. on Mobile and Ubiquitous Multimedia}, MUM '18, page 7–18, 2018.

\bibitem[Serpiva et~al.(2024)Serpiva, Fedoseev, Karaf, Abdulkarim, and Tsetserukou]{omnirace6dhandpose}
V.~Serpiva, A.~Fedoseev, S.~Karaf, A.~A. Abdulkarim, and D.~Tsetserukou.
\newblock Omnirace: 6d hand pose estimation for intuitive guidance of racing drone.
\newblock In \emph{IEEE/RSJ Int. Conf. on Int. Robots and Systems (IROS)}, pages 2508--2513, 2024.

\bibitem[Mellet et~al.(2024)Mellet, Berra, Marcellini, Ángel Trujillo~Soto, Heredia, Ruggiero, and Lippiello]{mellet2024designcontrolomnidirectionalaerial}
J.~Mellet, A.~Berra, S.~Marcellini, M.~Ángel Trujillo~Soto, G.~Heredia, F.~Ruggiero, and V.~Lippiello.
\newblock Design and control of an omnidirectional aerial robot with a miniaturized haptic joystick for physical interaction.
\newblock 2024.
\newblock arXiv:2410.09003.

\bibitem[Jongbloed et~al.(2024)Jongbloed, Chaker, and Lavou\'{e}]{Immersivetraining}
J.~Jongbloed, R.~Chaker, and E.~Lavou\'{e}.
\newblock Immersive procedural training in virtual reality: A systematic literature review.
\newblock \emph{Comput. Educ.}, 221\penalty0 (C), Nov. 2024.

\bibitem[Klingenberg et~al.(2024)Klingenberg, Bosse, Mayer, et~al.]{klingenberg2024does}
S.~Klingenberg, R.~Bosse, R.~E. Mayer, et~al.
\newblock Does embodiment in virtual reality boost learning transfer? testing an immersion-interactivity framework.
\newblock \emph{Educational Psychology Review}, 36:\penalty0 116, 2024.

\end{thebibliography}

\end{document}